\documentclass[runningheads]{llncs}
\usepackage{graphicx}

\usepackage{tikz}
\usepackage{comment}
\usepackage{amsmath,amssymb} 
\usepackage{color}
\usepackage{booktabs}
\usepackage{multirow}
\usepackage{bbding}
\usepackage{pifont}
\usepackage{wasysym}
\usepackage{amssymb}
\usepackage{pifont}

\usepackage{indentfirst}

\usepackage[accsupp]{axessibility}  

\begin{document}
\pagestyle{headings}
\mainmatter
\def\ECCVSubNumber{2218}  

\title{Teaching Where to Look: Attention Similarity Knowledge Distillation for Low Resolution Face Recognition} 
\titlerunning{Teaching Where to Look: Attention Similarity Knowledge Distillation}

\author{Sungho Shin \inst{1} \orcidID{0000-0001-5393-6169} \and
        Joosoon Lee \inst{1}\orcidID{0000-0001-6262-5303} \and
        Junseok Lee \inst{1}\orcidID{0000-0001-5212-2657} \and
        Yeonguk Yu \inst{1}\orcidID{0000-0003-2147-4718} \and
        Kyoobin Lee \inst{1}\orcidID{0000-0003-4299-4923} \thanks{Corresponding author.}}

\institute{School of Integrated Technology (SIT), Gwangju Institute of Science and Technology (GIST), Cheomdan-gwagiro 123, Buk-gu, Gwangju 61005, Republic of Korea. \\
\email{\{hogili89,joosoon1111,junseoklee,yeon\_guk,kyoobinlee\}\\@gist.ac.kr}}
\authorrunning{S. Shin et al.}

\maketitle
\begin{abstract}
Deep learning has achieved outstanding performance for face recognition benchmarks, but performance reduces significantly for low resolution (LR) images. We propose an attention similarity knowledge distillation approach, which transfers attention maps obtained from a high resolution (HR) network as a teacher into an LR network as a student to boost LR recognition performance. Inspired by humans being able to approximate an object's region from an LR image based on prior knowledge obtained from HR images, we designed the knowledge distillation loss using the cosine similarity to make the student network's attention resemble the teacher network's attention. Experiments on various LR face related benchmarks confirmed the proposed method generally improved recognition performances on LR settings, outperforming state-of-the-art results by simply transferring well-constructed attention maps. The code and pretrained models are publicly available in the \url{https://github.com/gist-ailab/teaching-where-to-look}.

\keywords{Attention similarity knowledge distillation, cosine similarity, low resolution face recognition}
\end{abstract}

\section{Introduction}
Recent face recognition model recognizes the identity of a given face image
from the 1M distractors with an accuracy of 99.087\%~\cite{Kemelmacher-Shlizerman2015}. However, most face recognition benchmarks such as MegaFace~\cite{Kemelmacher-Shlizerman2015}, CASIA~\cite{Yi2014}, and~MS-Celeb-1M \cite{Guo2016} contain high resolution (HR) images that differ significantly from real-world environments, typically captured by surveillance cameras. When deep learning approaches are directly applied to low resolution (LR) images after being trained on HR images, significant performance degradation occurred~\cite{Cheng2018LowResolutionFR,10.5555/1736406.1736429,5634490}. 

\begin{figure}
\centering
\includegraphics[height=5.0cm]{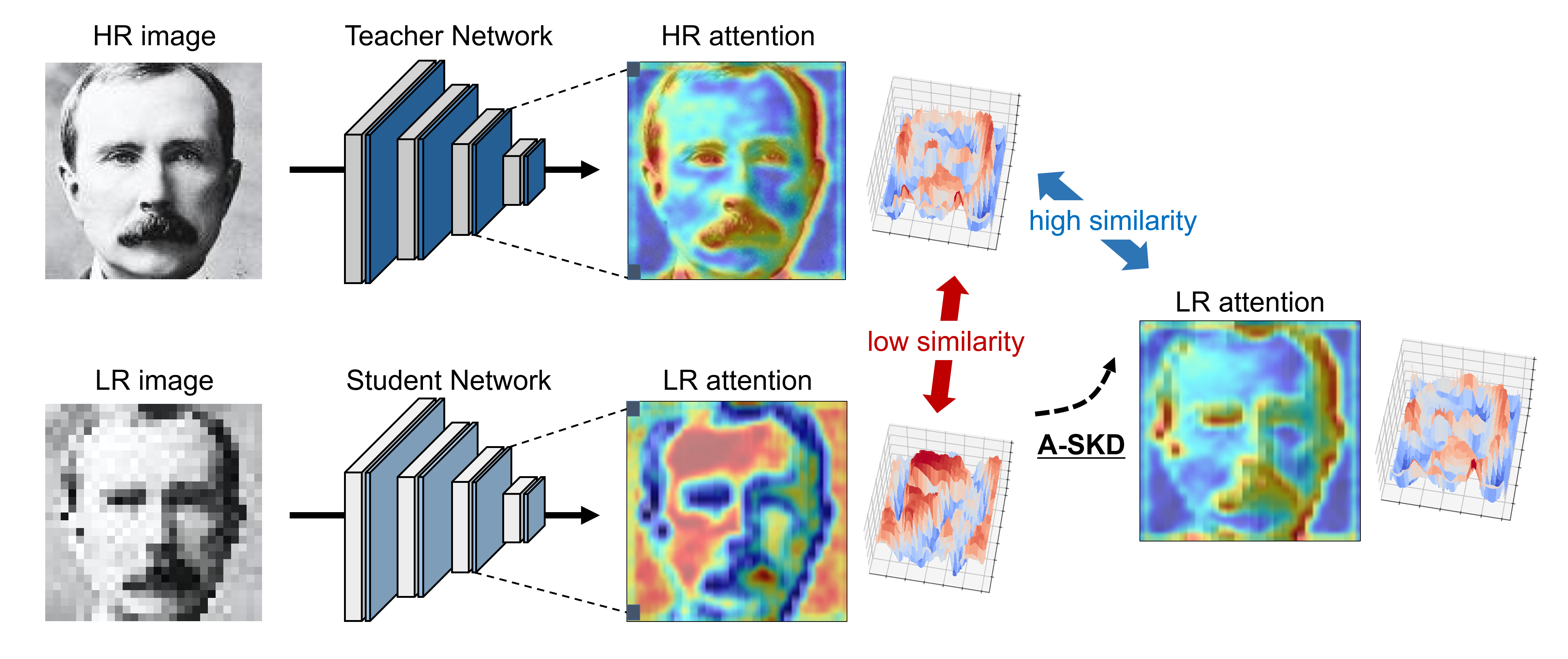}
\caption{Proposed attention similarity knowledge distillation (A-SKD) concept for low resolution (LR) face recognition problem. Well-constructed attention maps from the HR network are transferred to the LR network by forming high similarity between them for guiding the LR network to focus on detailed parts captured by the HR network. Face images and attention maps are from the AgeDB-30~\cite{inproceedings}.}
\label{fig:concept}
\end{figure}

To overcome the LR problem associated with face recognition, prior knowledge extracted from HR face images is used to compensate spatial information loss. Depending on the approach of transferring the prior knowledge to LR image domain, LR face recognition methods are categorized into two types: super-resolution and knowledge distillation based approaches. Super-resolution based approaches utilize generative models to improve LR images to HR before input to recognition networks~\cite{Fookes2012,Gunturk2003,Hennings-Yeomans2008,Kong2019,Tran2017,5634490}. Following the development of super-resolution methods, LR images can be successfully reconstructed into HR images and recognized by a network trained on HR images~\cite{Dong2016AcceleratingTS,7364266,Ledig2017PhotoRealisticSI,8889765}. However, super-resolution models incur high computational costs for both training and inference, even larger than the costs required for recognition networks. Furthermore, generating HR from LR images is an ill-posed problem, i.e., many HR images can match with a single LR image~\cite{srcnn}; hence the identity of a LR image can be altered.
 
To combat this, knowledge distillation based methods have been proposed to transfer prior knowledge from HR images to models trained on LR face images~\cite{Ge_Zhang_Liu_Hua_Zhao_Jin_Wen_2020,Massoli2020,8682926}. When the resolution of face images is degraded, face recognition models cannot capture accurate features for identification due to spatial information loss. In particular, features from detailed facial parts are difficult to be captured from a few pixels on LR images, e.g. eyes, nose, and mouth~\cite{S2LD}. Previous studies mainly focused on feature based knowledge distillation (F-KD) methods to encourage the LR network's features to mimic the HR network's features by reducing the Euclidean distance between them~\cite{Ge_Zhang_Liu_Hua_Zhao_Jin_Wen_2020,Massoli2020,8682926}. The original concept of F-KD was proposed as a lightweight student model to mimic features from over-parameterized teacher models~\cite{F-KD}. Because teacher model's features would generally include more information than the student model, F-KD approaches improve the accuracy of the student model. Similarly, informative features from the HR network are distilled to the LR network in the LR face recognition problems.

This study proposes the attention similarity knowledge distillation approach to distill well-constructed attention maps from an HR network into an LR network by increasing similarity between them. The approach was motivated by the observation that humans can approximate an object's regions from LR images based on prior knowledge learned from previously viewed HR images. Kumar et al. proposed that guiding the LR face recognition network to generate facial keypoints (e.g., eyes, ears, nose, and lips) improved recognition performance by directing the network's attention to the informative regions~\cite{S2LD}. Thus, we designed the prior knowledge as an attention map and transferred the knowledge by increasing similarity between the HR and LR networks' attention maps. 

Experiments on LR face recognition, face detection, and general object classification demonstrated that the attention mechanism was the best prior knowledge obtainable from the HR networks and similarity was the best method for transferring knowledge to the LR networks. Ablation studies and attention analyses demonstrated the proposed A-SKD effectiveness.

\section{Related Works}
\textbf{Knowledge distillation.} Hinton et al. first proposed the knowledge distillation approach to transfer knowledge from a teacher network into a smaller student network~\cite{KD}. Soft logits from a teacher network were distilled into a student network by reducing the Kullback-Leibler (KL) divergence score, which quantifies the difference between the teacher and student logits distributions. Various F-KD methods were subsequently proposed to distill intermediate representations~\cite{Park2019,FitNet,F-KD,AT}. FitNet reduced the Euclidean distance between teacher and student network's features to boost student network training~\cite{FitNet}. Zagoruyko et al. proposed attention transfer (AT) to reduce the distance between teacher and student network's attention maps rather than distilling entire features~\cite{AT}. Since attention maps are calculated by applying channel-wise pooling to feature vectors, activation levels for each feature can be distilled efficiently. Relational knowledge distillation (RKD) recently confirmed significant performance gain by distilling structural relationships for features across teacher and student networks~\cite{Park2019}. 
 
\textbf{Feature guided LR face recognition.} Various approaches that distill well-constructed features from the HR face recognition network to the LR network have been proposed to improve LR face recognition performances~\cite{Ge_Zhang_Liu_Hua_Zhao_Jin_Wen_2020,Massoli2020,8682926}. Conventional knowledge distillation methods assume that over-parameterized teacher networks extract richer information and it can be transferred to smaller student networks. Similarly, LR face recognition studies focused on transferring knowledge from networks trained on highly informative inputs to networks trained on less informative inputs. Zhu et al. introduced knowledge distillation approach for LR object classification \cite{8682926}, confirming that simple logit distillation from the HR to LR network significantly improved LR classification performance, even superior to super-resolution based methods. F-KD~\cite{Massoli2020} and hybrid order relational knowledge distillation (HORKD)~\cite{Ge_Zhang_Liu_Hua_Zhao_Jin_Wen_2020}, which is the variant of RKD~\cite{Park2019}, methods were subsequently applied to LR face recognition problems to transfer intermediate representations from the HR network.

Another approach is to guide the LR network by training it to generate keypoints (e.g. eyes, ears, nose, and lips)~\cite{S2LD}. An auxiliary layer is added to generate keypoints, and hence guide the network to focus on specific facial characteristics. It is well known that facial parts such as eyes and ears are important for recognition~\cite{landmark,S2LD}, hence LR face recognition networks guided by keypoints achieve better performance. Inspired by this, we designed the attention distillation method that guides the LR network to focus on important regions of the HR network. However, attention distillation methods have not been previously explored for LR face recognition. We investigated the efficient attention distillation methods for LR settings and proposed the cosine similarity as the distance measure between HR and LR network's attention maps.

\section{Method}
\subsection{Low resolution image generation} 
\begin{figure}[h!]
    \centering
    \includegraphics[scale=0.43]{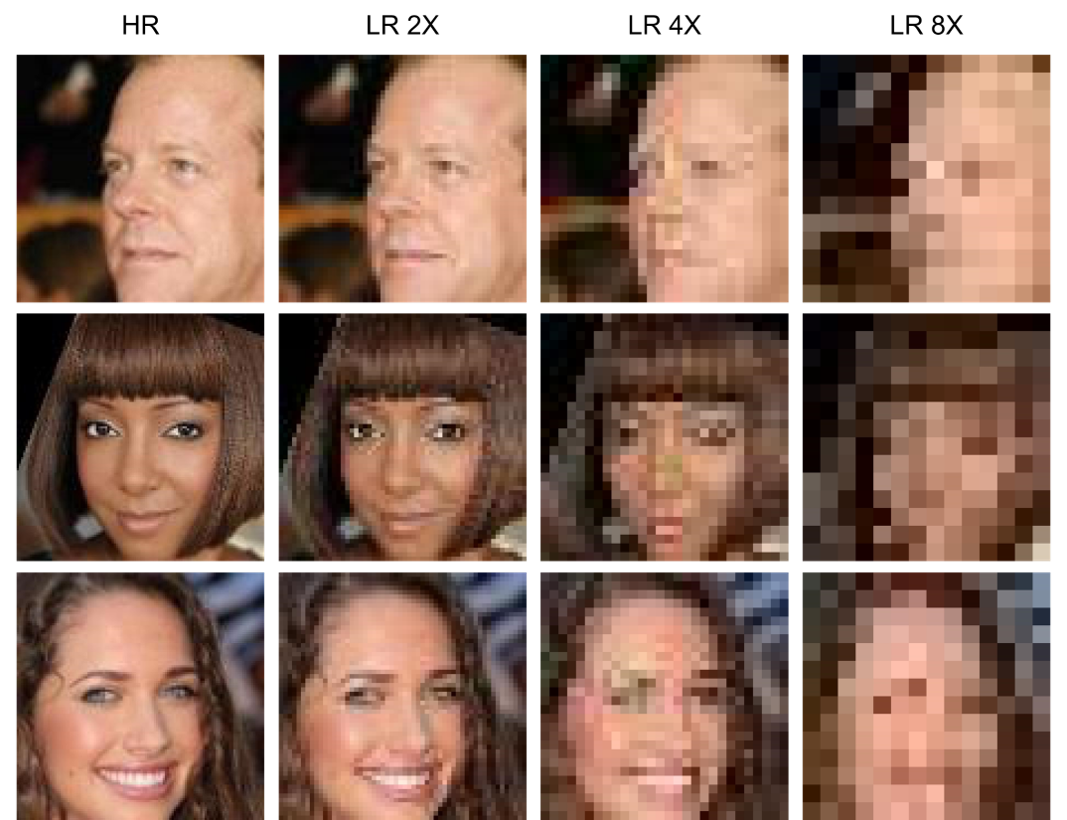}
    \caption{The samples of HR and LR images from the training dataset (CASIA \cite{Yi2014}) with the down-sampling ratios of 2$\times$, 4$\times$, and 8$\times$.}
\label{fig:face_images}
\end{figure}

We require HR and LR face image pairs to distill the HR network's knowledge to the LR network. Following the protocol for LR image generation in super-resolution studies~\cite{Dong2016AcceleratingTS,7364266,Ledig2017PhotoRealisticSI,8889765}, we applied bicubic interpolation to down-sample HR images with 2$\times$, 4$\times$, and 8$\times$ ratios. Gaussian blur was then added to generate realistic LR images. Finally, the downsized images were resized to the original image size using bicubic interpolation. Figure \ref{fig:face_images} presents sample LR images.

\subsection{Face recognition with attention modules}
\textbf{Face recognition network.} ArcFace \cite{Deng2018} is a SOTA face recognition network comprising convolutional neural network (CNN) backbone and angular margin introduced to softmax loss. Conventional softmax loss can be expressed as
\begin{equation} 
\label{eq:softmax}
L_{softmax} = -\frac{1}{N} \sum_{i=1}^{N}log\frac{e^{W^T_{y_i}x_i + b_{y_i}}}{\sum_{j=1}^{n}e^{W^T_jx_i + b_j}} , 
\end{equation}
where $x_i \in \mathbb{R}^d$ is the embedded feature of the $i$-th sample belonging to the $y_i$-th class; $N$ and $n$ are the batch size and the number of classes, respectively; $W_j \in \mathbb{R}^d$ denotes the $j$-th column of the last fully connected layer's weight $W \in \mathbb{R}^{d \times n}$ and $b_j \in \mathbb{R}^n$ is the bias term for the $j$-th class. 

For simplicity, the bias term is fixed to 0 as in \cite{sphereface}. Then the logit of the $j$-th class can be represented as $W^T_j x_i = \lVert{W_j}\rVert \lVert{x_i}\rVert {cos(\theta_j)}$, where $\theta_j$ denotes the angle between the $W_j$ and $x_i$. Following previous approaches~\cite{sphereface,cosface}, ArcFace set $\lVert{W_j}\rVert=1$ and $\lVert{x_i}\rVert=1$ via $l_2$ normalisation to maximize $\theta_j$ among inter-class and minimize $\theta_j$ among intra-class samples. Further, constant linear angular margin ($m$) was introduced to avoid convergence difficulty. The ArcFace~\cite{Deng2018} loss can be expressed as

\begin{equation}
\label{eq:arcface}
L_{arcface} = -\frac{1}{N} \sum_{i=1}^{N}log\frac{e^{s(cos(\theta_{y_i} + m))}}{e^{s(cos(\theta_{y_i} + m)) + \sum_{j=1,j\ne {y_i}}^{n}e^{s(cos(\theta_j))}}} ,
\end{equation}
where $s$ is the re-scale factor and $m$ is the additive angular margin penalty between $x_i$ and $W_{y_i}$.

\textbf{Attention.} Attention is a simple and effective method to guide feature focus on important regions for recognition. Let $\mathbf{f}_i = \mathcal{H}_i(\mathbf{x})$ be intermediate feature outputs from the \textit{i}-th layer of the CNN. Attention maps about $\mathbf{f}_i$ can be represented as the $\mathcal{A}_i(\mathbf{f}_i)$, where $\mathcal{A}_i(\cdot)$ is attention module. 

Many attention mechanisms have been proposed; AT \cite{AT} simply applied channel-wise pooling to features to estimate spatial attention maps. SENet~\cite{senet} and CBAM \cite{Woo2018} utilized parametric transformations, e.g. convolution layers, to represent attention maps. Estimated attention maps were multiplied with the features and passed to a successive layer. Trainable parameters in attention module are updated to improve performance during back-propagation, forming accurate attention maps. Attention mechanisms can be expressed as 
\begin{equation}
\label{eq:refine1}
\mathbf{f^{'}_{i}} = \mathcal{A}^c_{i}(\mathbf{f_{i}}) \otimes \mathbf{f_{i}}
\end{equation}
and
\begin{equation}
\label{eq:refine2}
\mathbf{f^{''}_{i}} = \mathcal{A}^s_{i}(\mathbf{f^{'}_{i}}) \otimes \mathbf{f^{'}_{i}} ,
\end{equation}
where $\mathcal{A}^c_{i}(\cdot)$ and $\mathcal{A}^s_{i}(\cdot)$ are attention modules for channel and spatial attention maps, respectively.

\begin{figure}
\centering
\includegraphics[height=4.5cm]{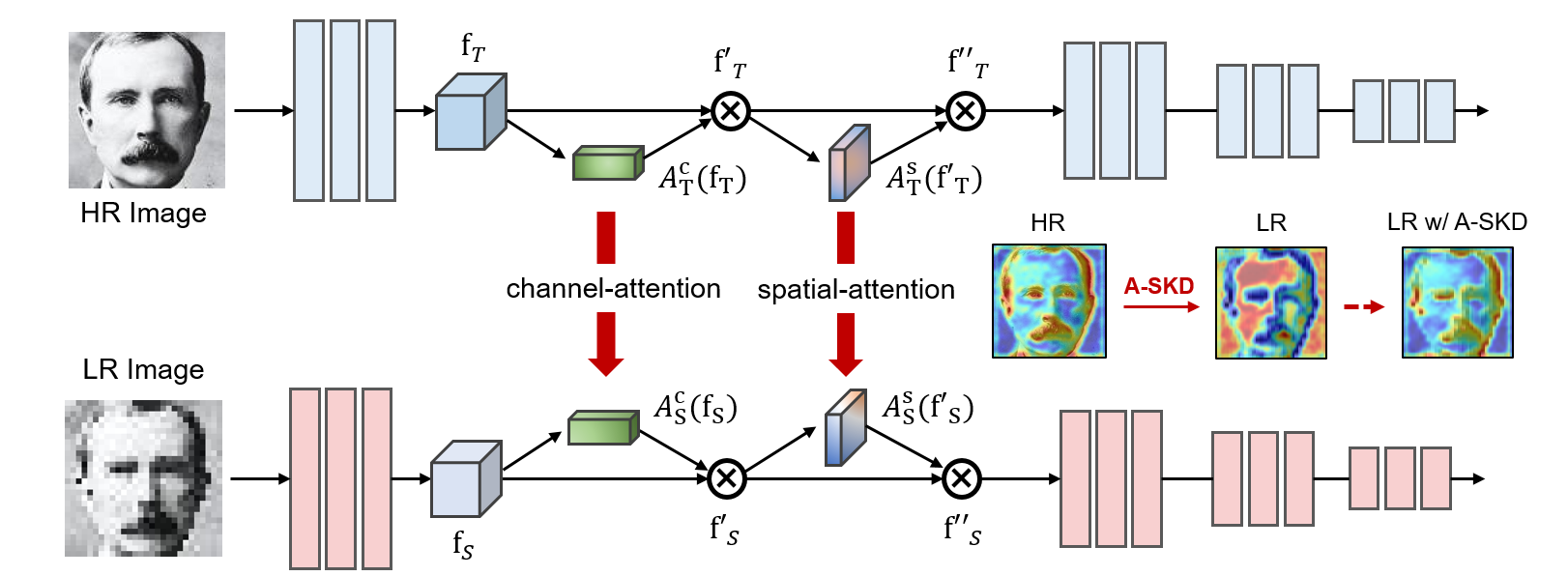}
\caption{Proposed A-SKD framework. The LR network formulates precise attention maps by referencing well-constructed channel and spatial attention maps obtained from the HR network, focusing on detailed facial parts which are helpful for the face recognition. We only show the attention distillation for the first block.}
\label{fig:model}
\end{figure}

Features are refined twice by multiplying channel and spatial attention maps in order (\ref{eq:refine1}) and (\ref{eq:refine2}). Any parametric attention transformation could be employed for the proposed A-SKD, and we adopted the popular CBAM~\cite{Woo2018} module, 
\begin{equation}
\label{eq:cam}
\mathcal{A}^c(\mathbf{f}) = \sigma (FC(AvgPool(\mathbf{f}))+FC(MaxPool(\mathbf{f})))
\end{equation}
and
\begin{equation}
\label{eq:sam}
\mathcal{A}^s(\mathbf{f}) = \sigma (f^{7\times7}(AvgPool(\mathbf{f});MaxPool(\mathbf{f}))) , 
\end{equation}
where $\sigma(\cdot)$ is the sigmoid function; and $FC(\cdot)$ and $f^{7\times7}(\cdot)$ are fully connected and convolution layers with $7\times7$ filters, respectively.

\subsection{Proposed attention similarity knowledge distillation framework}
Unlike the conventional knowledge distillation, the network size of teacher and student network is same for A-SKD. Instead, the teacher network is trained on HR images whereas the student network is trained on LR images. Due to the resolution differences, features from both networks are difficult to be identical. Therefore, we propose to distill well-constructed attention maps from the HR network into the LR network instead of features.

\begin{equation}
\begin{split}
\rho_i &= 1 - \langle\mathcal{A}_{T,i}(\mathbf{f}_{T,i}), \mathcal{A}_{S,i}(\mathbf{f}_{S,i})\rangle \\
       &= 1 - \frac{\mathcal{A}_{T,i}(\mathbf{f}_{T,i})}{\lVert \mathcal{A}_{T,i}(\mathbf{f}_{T,i}) \rVert_2} \cdot
      \frac{\mathcal{A}_{S,i}(\mathbf{f}_{S,i})}{\lVert \mathcal{A}_{S,i}(\mathbf{f}_{S,i})\rVert_2} , 
\end{split}
\end{equation}
where $\rho_i$ is the cosine distance between attention maps from the \textit{i}-th layer of the teacher and student networks; $\langle \cdot, \cdot \rangle$ denotes the cosine similarity; $\lVert \cdot \rVert_2$ denotes L2-norm; $\mathcal{A}_i(\mathbf{f}_i)$ denotes the attention maps for the \textit{i}-th layer features; and $T$ and $S$ denote the teacher and student network, respectively. Thus, $\mathcal{A}_{T,i}(\mathbf{f}_{T,i})$ and $\mathcal{A}_{S,i}(\mathbf{f}_{S,i})$ are attention maps estimated from the \textit{i}-th layer of the teacher and student network's features, respectively. Reducing the cosine distance between HR and LR attention maps increases the similarity between them.

Distillation loss for A-SKD is calculated as
\begin{equation}
\label{eq:kd_loss}
    \mathcal{L}_{distill} = \sum^{N}_{i=1} \frac{(\rho^s_i + \rho^c_i)}{2}
\end{equation}
which average the cosine distance for channel and spatial attention maps between the HR and LR networks, and sums them across layers ($i=1,2,3,...,N$) of the backbone. $N$ is the number of layers utilized for the distillation.

Total loss for the LR face recognition network is the sum of target task's loss and distillation loss (\ref{eq:kd_loss}) weighted by the factor ($\lambda_{distill}$). In this work, we utilized the ArcFace loss (\ref{eq:arcface}) as a target task's loss. 

\begin{equation}
\label{eq:tot_loss}
    \mathcal{L}_{total} = \mathcal{L}_{arcface} + \lambda_{distill} * \mathcal{L}_{distill}. 
\end{equation}

Further, our method can be utilized in conjunction with the logit distillation by simply adding the logit distillation loss~\cite{KD} to our loss function (\ref{eq:tot_loss}). Since logit is the final output of the network, incorporating the logit distillation loss allows the LR network to make the same decision as the HR network based on the refined attention maps.

\section{Experiments}

\subsection{Settings}
\textbf{Datasets.} We employed the CASIA~\cite{Yi2014} dataset for training, which is a large face recognition benchmark comprising approximately 0.5M face images for 10K identities. Each sample in CASIA was down-sampled to construct the HR-LR paired face dataset. For the evaluation, the manually down-sampled face recognition benchmark (AgeDB-30~\cite{inproceedings}) and the popular LR face recognition benchmark (TinyFace~\cite{Cheng2018LowResolutionFR}) were employed. Since AgeDB-30 have similar resolution to CASIA, networks trained on down-sampled CASIA images were validated on AgeDB-30 down-sampled images with matching ratio. In contrast, the real-world LR benchmark (TinyFace) comprises face images with the resolution of 24$\times$24 in average when they are aligned. Therefore, they were validated using a network trained on CASIA images down-sampled to 24$\times$24 pixels.

\textbf{Task and metrics.} Face recognition was performed for two scenarios: face verification and identification. Face verification is where the network determines whether paired images are for the same person, i.e.,  1:1 comparison. To evaluate verification performance, accuracy was determined using validation sets constructed from probe and gallery set pairs following the LFW protocol~\cite{LFW}. Face identification is where the network recognize the identity of a probe image by measuring similarity against all gallery images, i.e., 1:N comparison. This study employed the smaller AgeDB-30 dataset for the face verification; and larger TinyFace dataset for the face identification.

\textbf{Comparison with other methods.} Typically, the distillation of intermediate representation is performed concurrently with the target task’s loss. Previous distillation methods in the experiments utilized the both face recognition and distillation loss, albeit face recognition loss of varying forms. In addition, some feature distillation approach reported their performances with the logit distillation loss. In order to conduct a fair comparison, we re-implemented the prior distillation methods with the same face recognition loss (ArcFace~\cite{Deng2018}) and without the logit distillation loss. Further, our method requires the parametric attention modules for the distillation. Therefore, we utilized the same backbone network with CBAM attention modules for all methods; we combined the CBAM modules to all convolution layers, with the exception of the stem convolution layer and the convolution layer with a kernel size of 1.

\textbf{Implementation details.}
We followed the ArcFace protocol for data preprocessing: detecting face regions using the MTCNN~\cite{Zhang2016} face detector, cropping around the face region, and resizing the resultant portion to 112$\times$112 pixel using bilinear interpolation. The backbone network was ResNet-50 with CBAM attention module. For the main experiments, we distilled the attention maps for every convolution layers with the exception of the stem convolution layer and the convolution layer with a kernel size of 1. Weight factors for distillation (\ref{eq:tot_loss}) $\lambda_{distill} = 5$. This weight factors generally achieved the superior results not only for the face recognition benchmarks, but also for the ImageNet~\cite{imagenet}. Learning rate = 0.1 initially, divided by 10 at 6, 11, 15, and 17 epochs. SGD optimizer was utilized for the training with batch size = 128. Training completed after 20 epochs. The baseline refers to the LR network that has not been subjected to any knowledge distillation methods. For the hyperparameter search, we divided 20\% of the training set into the validation set and conducted a random search. After the hyperparameter search, we trained the network using the whole training set and and performed the evaluation on the AgeDB-30 and TinyFace.

\subsection{Face recognition benchmark results}
\textbf{Evaluation on AgeDB-30.}
Table \ref{tab:performances} shows LR face recognition performance on AgeDB-30 with various down-sample ratios depending on distillation methods. Except for HORKD, previous distillation methods~\cite{Massoli2020,AT} exhibited only slight improvement or even reduced performance when the downsampling ratios increase. This indicates that reducing the L2 distance between the HR and LR network's features is ineffective. In contrast, HORKD improved LR recognition performance by distilling the relational knowledge of the HR network's features. When the input’s resolution decrease, the intermediate features are hard to be identical with the features from the HR network. Instead the relation among the features of the HR network can be transferred to the LR network despite the spatial information loss; this was the reason of HORKD’s superior performances even for the 4$\times$ and 8$\times$ settings.

\begin{table}[]
\centering
\caption{Proposed A-SKD approach compared with baseline and previous SOTA methods on AgeDB-30 with 2$\times$, 4$\times$, and 8$\times$ down-sampled ratios. L, F, SA, and CA indicate distillation types of logit, feature, spatial attention, and channel attention, respectively. Ver-ACC denotes the verification accuracy. Base refers to the LR network that has not been subjected to any knowledge distillation methods.}
\resizebox{0.82\columnwidth}{!}{%
\begin{tabular}{@{}ccccc@{}}
\toprule
    \multirow{2}{*}{\textbf{Resolution}} &
    \multirow{2}{*}{\textbf{Method}} &
    \multirow{2}{*}{\textbf{Distill Type}} &
    \multirow{2}{*}{\textbf{Loss Function}} &
    \multirow{2}{*}{\textbf{\begin{tabular}[c]{@{}c@{}}Ver-ACC (\%)\\ (AgeDB-30)\end{tabular}}} \\
    & & & & \\ \midrule
    
                1$\times$  & Base        & -     & -          & 93.78 \\ \hline
\multirow{7}{*}{2$\times$} & Base        & -     & -          & 92.83 \\
                    & F-KD \cite{Massoli2020}        & F     & L2          &  93.05    \\
                    & AT \cite{AT}          & SA    & L2          & 92.93   \\
                    & HORKD \cite{Ge_Zhang_Liu_Hua_Zhao_Jin_Wen_2020}    & F     & L1+Huber  & 93.13   \\
                    & A-SKD \textbf{(Ours)} & SA+CA & Cosine           & \textbf{93.35}   \\ 
                    & A-SKD+KD \textbf{(Ours)} & SA+CA+L & Cosine+KLdiv    & \textbf{93.58}   \\   \hline
\multirow{7}{*}{4$\times$} & Base        & -     & -          &  87.74 \\
                    & F-KD \cite{Massoli2020}        & F     & L2         &  87.72    \\
                    & AT \cite{AT}          & SA    & L2          & 87.75   \\
                    & HORKD \cite{Ge_Zhang_Liu_Hua_Zhao_Jin_Wen_2020}         & F     & L1+Huber   & 88.08   \\
                    & A-SKD \textbf{(Ours)} & SA+CA & Cosine    & \textbf{88.58}   \\
                    & A-SKD+KD \textbf{(Ours)} & SA+CA+L & Cosine+KLdiv    & \textbf{89.15}   \\   \hline
\multirow{7}{*}{8$\times$} & Base        & -     & -          & 77.75 \\
                    & F-KD \cite{Massoli2020}        & F     & L2          &  77.85    \\
                    & AT \cite{AT}         & SA    & L2         & 77.40   \\
                    & HORKD \cite{Ge_Zhang_Liu_Hua_Zhao_Jin_Wen_2020}         & F     & L1+Huber   & 78.27   \\
                    & A-SKD \textbf{(Ours)} & SA+CA & Cosine         & \textbf{79.00}   \\ 
                    & A-SKD+KD \textbf{(Ours)} & SA+CA+L & Cosine+KLdiv    & \textbf{79.45}   \\ \bottomrule
\end{tabular}%
}
\label{tab:performances}
\end{table}

However, attention maps from the HORKD exhibit similar pattern to LR baseline network rather than the HR network in the Figure~\ref{fig:attn_example}. HR attention maps are highly activated in facial landmarks, such as eyes, lips, and beard, which are helpful features for face recognition~\cite{S2LD}. In contrast, detailed facial parts are less activated for LR attention maps because those parts are represented with a few pixels. Although HORKD boosts LR recognition performance by transferring HR relational knowledge, it still failed to capture detailed facial features crucial for recognition. The proposed A-SKD method directs the LR network's attention toward detailed facial parts that are well represented by the HR network's attention maps.

Based on the refined attention maps, A-SKD outperforms the HORKD and other knowledge distillation methods for all cases. AgeDB-30 verification accuracy increased 0.6\%, 1.0\%, and 1.6\% compared with baseline for 2$\times$, 4$\times$, and 8$\times$ down-resolution ratios, respectively. In addition, when A-SKD is combined with logit distillation (KD), the verification accuracy increased significantly for all settings. From the results, we confirmed that the attention knowledge from the HR network can be transferred to the LR network and led to significant improvements that were superior to the previous SOTA method.

\textbf{Evaluation on TinyFace.} Unlike the face verification, the identification task requires to select a target person's image from the gallery set consists of a large number of face images. Therefore, the identification performances decrease significantly when the resolution of face images are degraded. Table \ref{tab:tinyface} showed the identification performances on the TinyFace benchmark. When the AT~\cite{AT} was applied, the rank-1 identification accuracy decreased 13.34\% compared to the baseline. However, our approach improved the rank-1 accuracy 13.56\% compared to the baseline, even outperforming the HORKD method. This demonstrated that the parametric attention modules (CBAM) and cosine similarity loss are the key factors for transferring the HR network's knowledge into the LR network via attention maps. The proposed method is generalized well to real-world LR face identification task which is not manually down-sampled.

\begin{table}[]
\centering
\caption{Evaluation results on TinyFace identification benchmark depending on the distillation methods. Acc@K denotes the rank-K accuracy (\%).}
\label{tab:tinyface}
\resizebox{0.60\textwidth}{!}{%
\begin{tabular}{@{}ccccc@{}}
\toprule
\textbf{} & \textbf{ACC@1} & \textbf{ACC@5} & \textbf{ACC@10} & \textbf{ACC@50} \\ \midrule
Base      &     42.19       &       50.62         &         53.67        &          60.41       \\
AT~\cite{AT}        &       36.56       &       45.68        &   49.03   &  56.44               \\
HORKD~\cite{Ge_Zhang_Liu_Hua_Zhao_Jin_Wen_2020}     &      45.49       &    54.80       &   58.26    &  64.30   \\
A-SKD \textbf{(Ours)}     &  \textbf{47.91}    &    \textbf{56.55}     &    \textbf{59.92}   &   \textbf{66.60} \\ \bottomrule
\end{tabular}%
}
\end{table}

\section{Discussion}
\begin{figure}
\centering
\includegraphics[height=5.1cm]{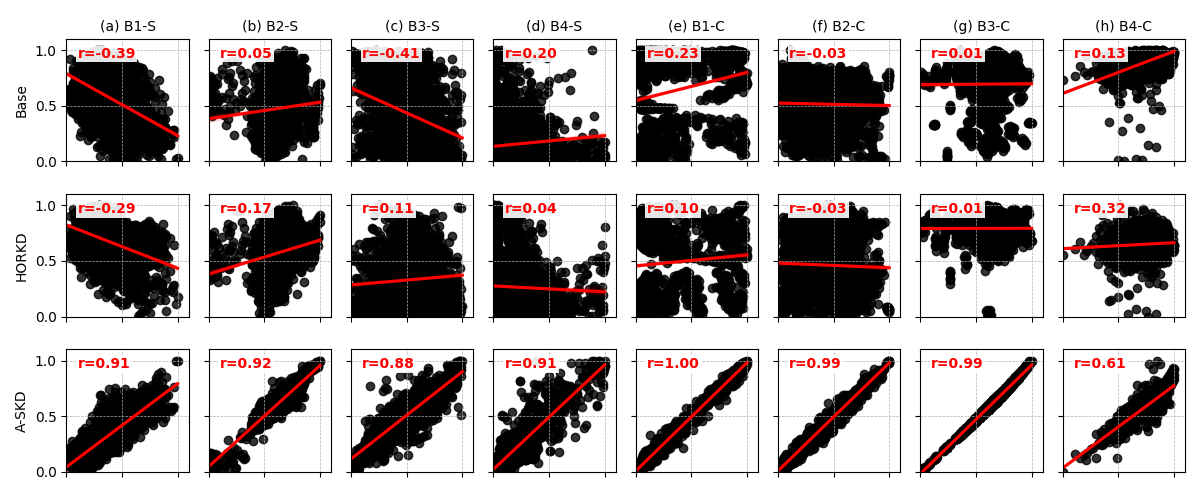}
\caption{Pixel level Pearsons correlation between the HR and LR network's attention maps for different distillation methods. B\{$i$\}-\{S,C\} indicates Pearsons correlation for spatial or channel attention maps obtained from the $i$-th ResNet block between the HR and LR networks; and $r$ is Pearsons correlation coefficient representing linear relationships between input variables. Base refers to the LR network that has not been subjected to any knowledge distillation methods. Pearsons correlation is measured using the AgeDB-30.}
\label{fig:correlation}
\end{figure}

\textbf{Attention correlation analysis.} Figure \ref{fig:correlation} shows Pearsons correlation between attention maps from the HR and LR networks for the different distillation methods. Spatial and channel attention maps from the four blocks for models other than A-SKD have a low correlation between the HR and LR networks, with a magnitude lower than 0.5. In particular, spatial attention maps obtained from the first block of the LR baseline and HORKD network have negative correlation with the HR network ($r = -0.39$ and $-0.29$, respectively).

\begin{figure}
\centering
\includegraphics[height=7.3cm]{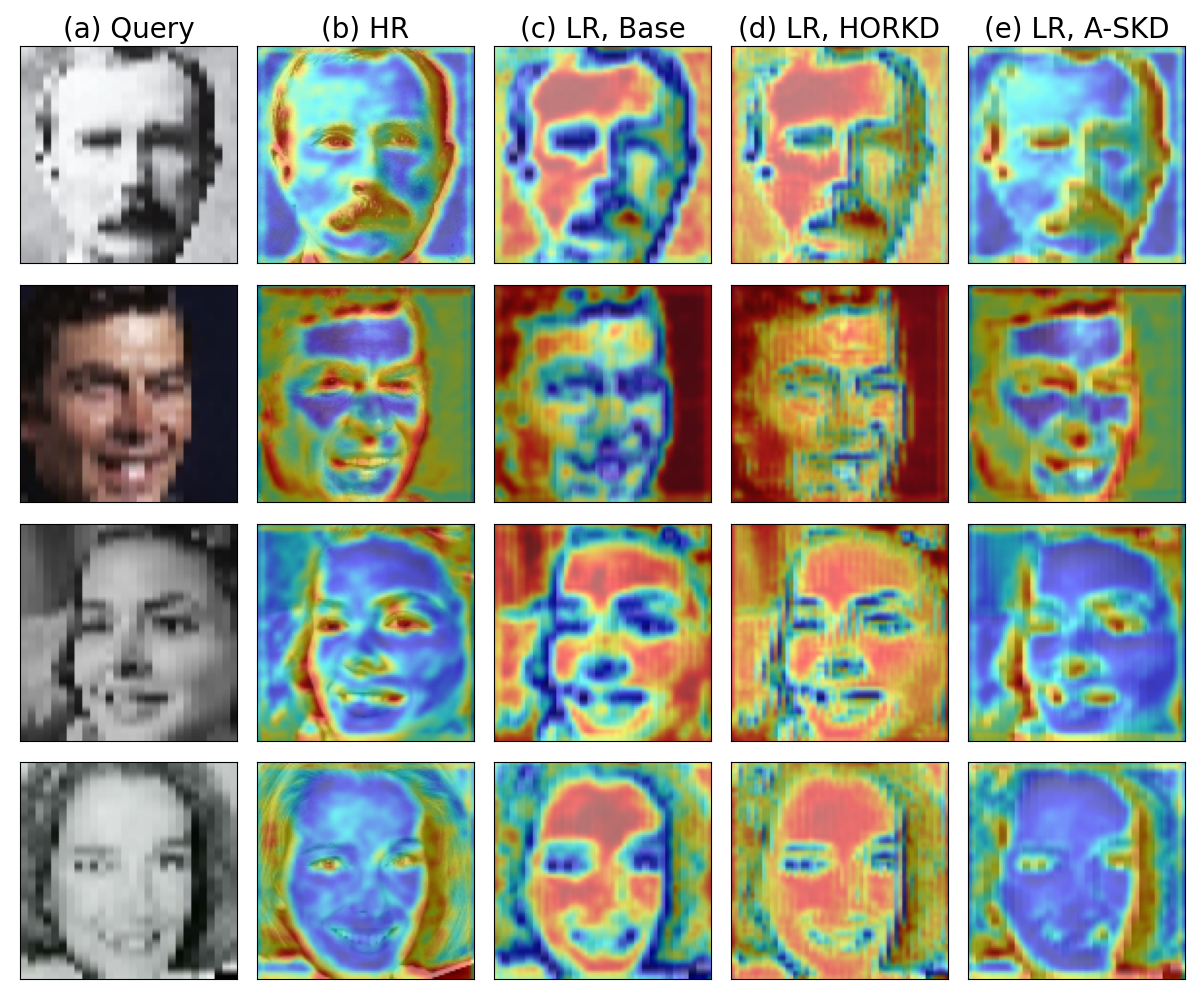}
\caption{Normalized spatial attention maps from the first block for different distillation methods. Red and blue regions indicate high and low attention, respectively. Face images and attention maps are from the AgeDB-30.}
\label{fig:attn_example}
\end{figure}

Figure \ref{fig:attn_example} shows that spatial attention maps from LR baseline and HORKD networks are highly activated in skin regions, which are less influenced by resolution degradation, in contrast to the HR network. This guides the LR network to the opposite directions from the HR network. However, spatial attention maps from A-SKD exhibit strong positive correlation with those from the HR network, highlighting detailed facial attributes such as beard, hair, and eyes. Through the A-SKD, the LR network learned where to focus by generating precise attention maps similar to those for the HR network. Consequently, Pearsons correlation, i.e., the similarity measure between HR and LR attention maps, was significantly improved for all blocks, with a magnitude higher than 0.6. Thus the proposed A-SKD approach achieved superior efficacy and success compared with previous feature based SOTA methods.

\textbf{Comparison with attention transfer~\cite{AT}.} Primary distinctions between AT \cite{AT} and A-SKD include the cosine similarity loss, parametric attention modules, and distillation of both channel and spatial attention maps. Correlation analysis for A-SKD confirmed that the cosine similarity loss is an effective strategy for transferring attention knowledge. Distilling AT attention maps using the cosine similarity rather than the L2 loss increased AgeDB-30 verification accuracy by 0.32\%p (Table~\ref{tab:ablation}). AT calculates attention maps using channel-wise pooling, a non-parametric layer; whereas A-SKD calculates attention maps using parametric layers comprising fully connected and convolution layers. When the input image resolution degrades, the student network’s feature representation diverges from that of the teacher network. Therefore, it is difficult to match the attention maps of the student network obtained by the non-parametric module with those of the teacher network. Instead, A-SKD employs the parametric module for the attention maps extraction and the cosine similarity loss for the distillation; therefore, the attention maps from the student network can be adaptively trained to be similar to the attention maps from the teacher network despite the differences in the features. Finally, A-SKD distills both spatial and channel attention maps in contrast to AT which only considered spatial attention maps. We confirmed A-SKD with spatial and channel attention additionally improved AgeDB-30 verification accuracy by 0.34\%p compared with spatial-only attention. This comparison results also confirmed that A-SKD, designed for attention distillation on LR settings, is the most effective approach for transferring attention knowledge.

\begin{table}[]
\centering
\caption{Comparing attention transfer (AT) \cite{AT} and proposed A-SKD on AgeDB-30 benchmark down-sampled with 8$\times$ ratio. AT* indicates the cosine similarity loss was utilized for attention transfer rather than the original L2 loss. SA and CA indicate spatial and channel attention maps, respectively.}
\resizebox{0.80\textwidth}{!}{%
\begin{tabular}{@{}ccccc@{}}
\toprule
\textbf{Method} &
  \textbf{Type} &
  \textbf{Transformation} &
  \textbf{Loss Function} &
  \textbf{\begin{tabular}[c]{@{}c@{}}Ver-ACC (\%)\\ (AgeDB-30)\end{tabular}} \\ \midrule
AT \cite{AT}    & SA & Non-parametric layer     & L2             & 77.40          \\
AT$^*$       & SA  & Non-parametric layer    & Cosine    & 77.72          \\
A-SKD & SA  & Parametric layer    & Cosine          & 78.66          \\
A-SKD & SA + CA & Parametric layer  & Cosine          & \textbf{79.00} \\ \bottomrule
\end{tabular}
\label{tab:ablation}%
}
\end{table}

\section{Extension to Other Tasks}
\subsection{Object classification}
We conducted experiments for object classification on LR images using the 4$\times$ down-sampled ImageNet~\cite{imagenet}. For the backbone network, we utilized the ResNet18 with CBAM attention modules. We compared our method to other knowledge distillation methods (AT~\cite{AT} and RKD~\cite{Park2019}) which are widely utilized in the classification domains. We re-implemented those methods using its original hyperparameters. Usually, AT and RKD were utilized along with the logit distillation for the ImageNet; therefore, we performed the AT, RKD, and A-SKD in conjunction with the logit distillation in the Table \ref{tab:imagenet_result}. Training details are provided in the Supplementary Information.

Table \ref{tab:imagenet_result} shows that A-SKD outperformed the other methods on the LR ImageNet classification task. Park et al. demonstrated that introducing the accurate attention maps led the significant improvement on classification performances~\cite{Park2018BAMBA,Woo2018}. When the attention maps were distilled from the teacher network, student network could focus on informative regions by forming precise attention maps similar with the teacher's one. Thus, our method can be generalized to general object classification task, not restricted to face related tasks.

\begin{table}[]
\centering
\caption{Proposed A-SKD performance on low resolution ImageNet classification. All distillation methods were performed in conjunction with the logit distillation.}
\label{tab:imagenet_result}
\resizebox{0.36\textwidth}{!}{%
\begin{tabular}{@{}ccc@{}}
\toprule
\textbf{Resolution} & \textbf{Method} & \textbf{ACC (\%)} \\ \midrule
1$\times$                  & Base            &      70.13        \\ \hline
\multirow{4}{*}{4$\times$} & Base            &      65.34        \\
                    & AT~\cite{AT}           &      65.79             \\
                    & RKD~\cite{Park2019}    &      65.95            \\
                    & A-SKD                  &      \textbf{66.52}      \\ \bottomrule
\end{tabular}%
}
\end{table}

\subsection{Face detection}
Face detection is a sub-task of object detection to recognize human faces in an image and estimate their location(s). We utilized TinaFace~\cite{TinaFace}, a deep learning face detection model, integrated with the CBAM attention module to extend the proposed A-SKD approach to face detection. Experiments were conducted on the WIDER FACE~\cite{7780965} dataset (32,203 images containing 393,703 faces captured from real-world environments) with images categorized on face detection difficulty: easy, medium, and hard. LR images were generated with 16$\times$ and 32$\times$ down-resolution ratios, and further training and distillation details are provided in the Supplementary Information.

\begin{table}[]
\centering
\caption{Proposed A-SKD performance on LR face detection. mAP is mean average precision; easy, medium, and hard are pre-assessed detection difficulty.}
\resizebox{0.46\textwidth}{!}{%
\begin{tabular}{ccccc}
\toprule
\multirow{2}{*}{\textbf{Resolution}} & \multirow{2}{*}{\textbf{Model}} & \multicolumn{3}{c}{\textbf{mAP (\%)}} \\ \cline{3-5} 
                     &                  & \textbf{Easy} & \textbf{Medium} & \textbf{Hard} \\ \midrule
1$\times$                   & Base & 95.56         & 95.07           & 91.45         \\ \hline
\multirow{2}{*}{16$\times$}  & Base & 54.38         & 52.73           & 35.29         \\
                     & A-SKD & \textbf{62.93}         & \textbf{60.19}           & \textbf{47.28}         \\ \hline
\multirow{2}{*}{32$\times$} & Base & 31.15         & 26.68           & 14.00         \\
                     & A-SKD & \textbf{33.50}         & \textbf{30.04}           & \textbf{16.02}         \\ \bottomrule
\end{tabular}%
}
\label{tab:detection-result}
\end{table}

\begin{figure}
\centering
\includegraphics[height=5.0cm]{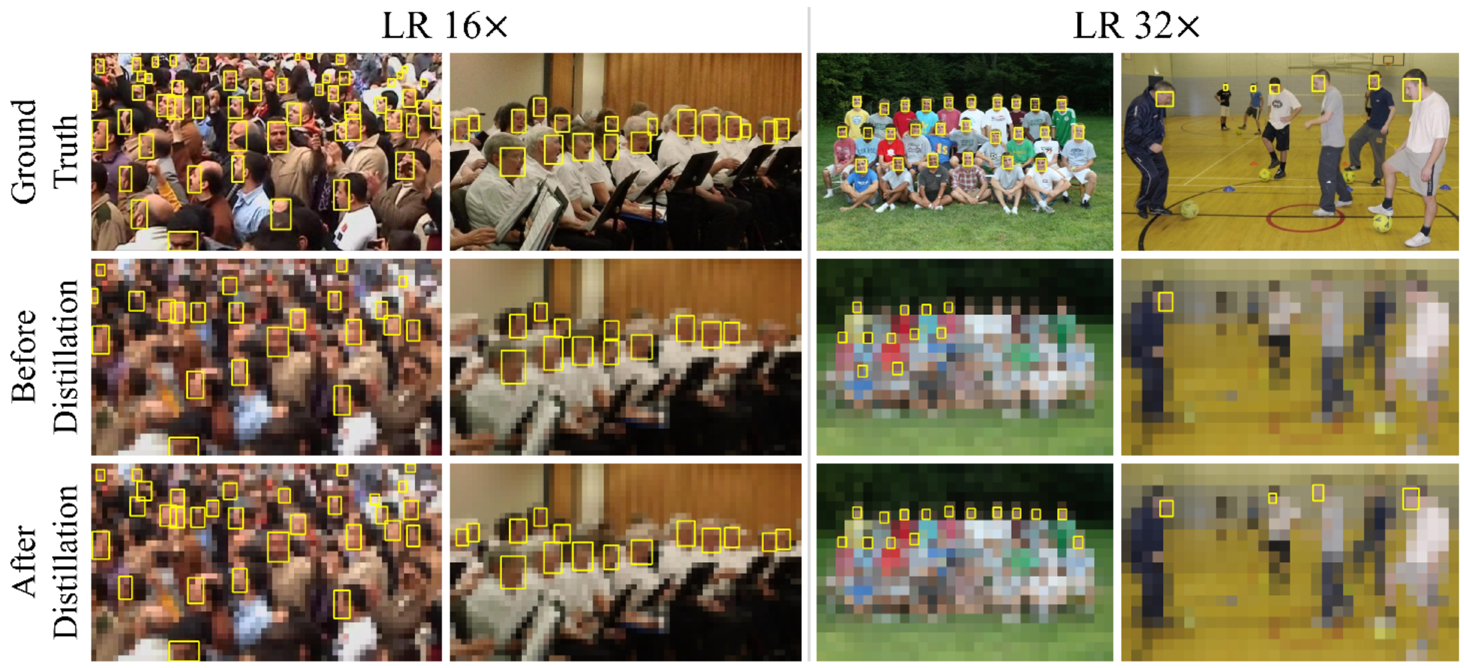}
\caption{Qualitative results for LR face detection before and after applying A-SKD. Small faces were better detected after A-SKD. The face images are from the evaluation set of WIDERFACE.}
\label{fig:detection_result}
\end{figure}

Table~\ref{tab:detection-result} shows that A-SKD improved the overall detection performance by distilling well-constructed attention maps, providing significant increases of mean average precision (mAP) for the easy (15.72\% for 16$\times$ and 7.54\% for 32$\times$), medium (14.15\% for 16$\times$ and 12.59\% for 32$\times$), and hard (33.98\% for 16$\times$ and 14.43\% for 32$\times$) level detection tasks. Small faces were well detected in the LR images after distillation as illustrated in Figure \ref{fig:detection_result}. Thus the proposed A-SKD approach can be successfully employed for many LR machine vision tasks.

\section{Conclusion}
We verified that attention maps constructed from HR images were simple and effective knowledge that can be transferred to LR recognition networks to compensate for spatial information loss. The proposed A-SKD framework enabled any student network to focus on target regions under LR circumstances and generalized well for various LR machine vision tasks by simply transferring well-constructed HR attention maps. Thus, A-SKD could replace conventional KD methods offering improved simplicity and efficiency and could be widely applicable to LR vision tasks, which have not been strongly studied previously, without being limited to face related tasks.  \\

\textbf{Acknowledgments}
This work was supported by the ICT R\&D program of MSIT/IITP[2020-0-00857, Development of Cloud Robot Intelligence Augmentation, Sharing and Framework Technology to Integrate and Enhance the Intelligence of Multiple Robots. And also, this work was partially supported by Korea Institute of Energy Technology Evaluation and Planning (KETEP) grant funded by the Korea government (MOTIE)(No. 20202910100030) and supported by Electronics and Telecommunications Research Institute (ETRI) grant funded by the Korean government. [22ZR1100, A Study of Hyper-Connected Thinking Internet Technology by autonomous connecting, controlling and evolving ways].

\clearpage
%
%
\bibliographystyle{splncs04}
\bibliography{egbib}
\end{document}


\pagestyle{headings}
\mainmatter
\def\ECCVSubNumber{2218}  

\title{Teaching Where to Look: Attention Similarity Knowledge Distillation for Low Resolution Face Recognition} 
\titlerunning{Teaching Where to Look: Attention Similarity Knowledge Distillation}

\author{Sungho Shin \inst{1}  \and
        Joosoon Lee \inst{1} \and
        Junseok Lee \inst{1} \and
        Yeonguk Yu \inst{1} \and
        Kyoobin Lee \inst{1} \thanks{Corresponding author.}}

\institute{School of Integrated Technology (SIT), Gwangju Institute of Science and Technology (GIST), Cheomdan-gwagiro 123, Buk-gu, Gwangju 61005, Republic of Korea. \\
\email{\{hogili89,joosoon1111,junseoklee,yeon\_guk,kyoobinlee\}\\@gist.ac.kr}}
\authorrunning{S. Shin et al.}
\maketitle

\section{Implementation Details of Previous Methods}
Various knowledge distillation approaches have been proposed to transfer the teacher network's knowledge to student network. Usually, the loss term of the knowledge distillation methods can be defined as the summation of the target task's loss and the distillation loss. $L_{total} = L_{target} + L_{distill}$. To compare the A-SKD with the previous distillation methods (HORKD~\cite{Ge_Zhang_Liu_Hua_Zhao_Jin_Wen_2020}, F-KD~\cite{Massoli2020}, and AT~\cite{AT}), we re-implemented those methods using the official implementation code.

The distillation loss function of the AT~\cite{AT} is defined as the L2 distance between the teacher and student network's self-attention maps. For the ResNet50, the AT distills the self-attention maps on the four blocks. We performed the experiments using the official implementation of AT (\url{https://github.com/szagoruyko/attention-transfer}).

The distillation loss function of the F-KD~\cite{Massoli2020} is defined as the L2 distance between the teacher and student network's feature maps. The penultimate layer's features (output of the backbone network before input to the margin) are utilized for the distillation. We performed the experiments using the official implementation of F-KD (\url{https://github.com/fvmassoli/cross-resolution-face-recognition}).

The loss function of the previous SOTA method (HORKD~\cite{Ge_Zhang_Liu_Hua_Zhao_Jin_Wen_2020}) is defined as
\begin{equation}
\label{eq:horkd_loss}
    \mathcal{L}_{distill} = L_1 + \alpha L_2 + \beta L_3 + \gamma L_C
\end{equation}
where $L_1, L_2, L_3$, and $L_C$ indicate the individual-level, pair-level, triplet-level, and group-level knowledge distillation loss, respectively; the different order relational distillation losses are tuned by the factor of $\alpha, \beta, \gamma$ in order. In our implementation, $\alpha=0.1$, $\beta=0.2$, and $\gamma=0.1$ after the hyperparameter search using the same protocol with our method. The penultimate layer's features (output of the backbone network before input to the margin) are utilized for the distillation. Because HORKD has no official implementation code, we referenced the official implementation of RKD~\cite{Park2019} (\url{https://github.com/lenscloth/RKD}).

\section{Extension to Other Tasks}
\subsection{Object Classification.} We utilized the ResNet50-CBAM backbone for the ImageNet training. The model was trained by SGD optimizer for 90 epochs with learning rate = 0.4, batch size = 1024, momentum = 0.9, and weight decay = 0.0001. Mixed precision was applied for the efficient training. We referenced the code of \url{https://github.com/rwightman/pytorch-image-models} for the training and validation on ImageNet benchmark.

\subsection{Face Detection}
The WIDER FACE~\cite{7780965} was utilized for training and validation to extend A-SKD to face detection. The WIDER FACE contained 32,203 images from 61 event classes, totaling 393,703 faces. The WIDER FACE categorized images into three face detection difficulties: easy, medium, and hard, considering scale, occlusion, and pose. Face detection performance on WIDER FACE was evaluated by overall mean average precision (mAP), and easy, medium, and hard images were evaluated separately. We trained TinaFace~\cite{TinaFace} on WIDER FACE by combining it with the CBAM attention module. LR images with 16$\times$ and 32$\times$ down-sampling ratio were trained to validate A-SKD effectiveness. TinaFace used ResNet-50 backbone network; and the loss function combined face detection and attention distillation losses. For the detection task, we distilled the attention maps of the four ResNet blocks, not every convolution layer. The model was trained by SGD optimizer for 150 epochs with learning rate = 0.001, momentum = 0.9, and weight decay = 0.0005 on two Titan RTX (24GB) GPUs with batch size = 8. We used non-maximum suppression threshold = 0.4 and confidence level = 0.02. We referenced the code of \url{https://github.com/Media-Smart/vedadet} for the implementation of TinaFace.

\clearpage
%
%
\bibliographystyle{splncs04}
\bibliography{egbib}